%% file: main.tex
\documentclass{aifrontiers}

\usepackage{aifrontiers} 
\date{\today}

\usepackage[utf8]{inputenc} 
\usepackage[T1]{fontenc}    
\usepackage{hyperref}       
\usepackage{url}            
\usepackage{booktabs}       
\usepackage{amsfonts}       
\usepackage{nicefrac}       
\usepackage{microtype}      
\usepackage{xcolor}

\usepackage{graphicx}
\usepackage{amsmath}
\usepackage{amssymb}
\usepackage{makecell}
\usepackage{array}
\usepackage{algorithm}
\usepackage{algpseudocode}
\usepackage{bbm}
\usepackage{wrapfig}
\usepackage{caption}
\usepackage{multirow}

\usepackage{cleveref}
\usepackage[normalem]{ulem}   

\providecommand{\claude}[2]{#2}

\title{Multi-Rollout On-Policy Distillation \\ via Peer Successes and Failures}

\author{
    \textbf{Weichen Yu$^{1,2}$, Xiaomin Li$^{2}$, Yizhou Zhao$^{1}$, Xiaoze Liu$^{3}$,}\\
    \textbf{ Ruowang Zhang$^{3}$, Haixin Wang$^{2}$, Yinyi Luo$^{1}$, Chen Henry Wu$^{1}$, Gaurav Mittal$^{2}$, \\
    Matt Fredrikson$^{1}$, Yu Hu$^{2}$}\\
    {\normalsize $^{1}$Carnegie Mellon University \quad $^{2}$Microsoft \quad $^{3}$Purdue University}\\
    {\normalsize  \texttt{wyu3@andrew.cmu.edu,} \quad  \texttt{xiaominli@microsoft.com}}
}

\begin{document}

\begin{abstract}
    Large language models are often post-trained with sparse verifier rewards, which indicate whether a sampled trajectory succeeds but provide limited guidance about where reasoning succeeds or fails. On-policy distillation (OPD) offers denser token-level supervision by training on student-generated trajectories, yet existing methods typically distill each rollout independently and ignore the other attempts sampled for the same prompt. We introduce Multi-Rollout On-Policy Distillation (MOPD), a peer-conditioned distillation framework that uses the student’s local rollout group to construct more informative teacher signals. MOPD conditions the teacher on both successful and failed peer rollouts: successes provide positive evidence for valid reasoning patterns, while failures provide structured negative evidence about plausible mistakes to avoid. \claude{We study several peer-context constructions, including positive peer imitation, contrastive success–failure conditioning, and summary-based peer distillation.}{We study two peer-context constructions: positive peer imitation and contrastive success--failure conditioning.} Experiments on competitive programming, mathematical reasoning, scientific question answering, and tool-use benchmarks show that MOPD consistently improves over standard on-policy baselines. Further teacher-signal analysis shows that mixed success–failure contexts better align teacher scores with verifier rewards, \claude{suggesting that the gains arise from more faithful, instance-adaptive supervision}{indicating that the gains arise from more faithful, instance-adaptive supervision}. These results indicate that effective on-policy distillation should exploit the student’s multi-rollout trial-and-error behavior rather than treating rollouts as isolated samples. Code is available at \url{https://github.com/viviable/mopd_code}.

\end{abstract}
\maketitle

\input{text/intro}

\input{text/related_work_short}
\input{text/prelim}

\input{text/method}
\input{text/exp_1}
\input{text/exp}

\input{text/exp_3}

\input{text/conclusion}

\bibliographystyle{plainnat} 
\bibliography{custom,custom_new}

\newpage
\appendix

\input{text/appendix}


\end{document}

%% file: text/intro.tex
\section{Introduction}
Large language models (LLMs) are often post-trained with reinforcement learning from verifiable rewards, where sampled solutions are scored by answer checkers, unit tests, or task-specific verifiers \cite{guo2025deepseek,schulman2017proximal,shao2024deepseekmath}. Although effective, these rewards are usually sparse: they indicate whether an entire trajectory succeeds or fails, but provide little guidance about which tokens or reasoning steps caused the outcome \cite{yue2025doesrl,chan2024dense}. Distillation and self-distillation provide a complementary form of supervision by converting teacher or privileged-model judgments into dense token-level signals \cite{zelikman2022star, gulcehre2023rest, singh2023restem, chen2024self, yang2024self}. On-policy distillation further aligns this supervision with the student’s own behavior by training on trajectories sampled from the student policy \cite{agarwal2024policy, 2306.08543, 2402.03898}.

Despite this advantage, existing OPD and on-policy self-distillation methods underuse a key source of information already available during training: multiple rollouts generated for the same problem instance. In many pipelines, each rollout is distilled independently, so the teacher signal for one trajectory is computed without access to the other attempts sampled from the same student. This design discards the local structure of the rollout group. For reasoning-intensive tasks, such local structure is highly informative: successful rollouts reveal valid solution strategies, while failed rollouts expose plausible but incorrect reasoning paths, missing constraints, formatting errors, or execution mistakes. Treating rollouts independently prevents the teacher from comparing these alternatives and identifying where an unsuccessful trajectory diverges from a successful one.

We propose \textit{Multi-Rollout On-Policy Distillation (MOPD)}, a peer-conditioned distillation framework that explicitly uses the student’s rollout group to construct teacher signals. For each prompt, the student samples multiple on-policy trajectories, which are scored by a verifier and partitioned into successful and failed sets. MOPD then conditions the teacher on peer rollouts from the same problem instance when supervising a target trajectory. In this formulation, successful rollouts act as peer experts that provide positive evidence about valid reasoning patterns, while failed rollouts serve as structured negative evidence that identifies misleading solution paths the student should avoid.

The key insight is that \textbf{multi-rollout sampling creates a local trial-and-error space around each problem}. Instead of viewing each trajectory in isolation, the teacher can compare the current rollout against peer successes and failures, using this contrast to produce more targeted supervision. This turns OPD from independent imitation of individual trajectories into a comparative learning process: the teacher is no longer only a global expert, but also a local diagnostic model that can recognize instance-specific errors and distinguish superficially plausible failures from correct solutions. We instantiate this idea through two peer-context construction strategies: positive peer imitation and contrastive success--failure conditioning.

{Moreover, prior methods often assume that a self-teacher becomes a reliable source of supervision once it is conditioned on privileged information, such as ground-truth outcomes, verified successful answers, hints, or environment feedback. However, this assumption is usually evaluated only indirectly through downstream training performance. Moreover, directly asking whether the privileged teacher can recover the correct answer is not a reliable diagnostic: if the privileged context already contains answer-related information or verifier feedback, the self-teacher may appear accurate by exploiting shortcuts rather than by producing a faithful supervision signal over the student’s own trajectories.} \textbf{What matters for self distillation is not merely whether the self-teacher knows the answer, but whether its token-level or trajectory-level preferences are aligned with correctness among the rollouts the student actually generates.}

\begin{wrapfigure}{r}{0.5\textwidth}
    \centering
    \vspace{-1.5em}
    \includegraphics[width=0.5\textwidth]{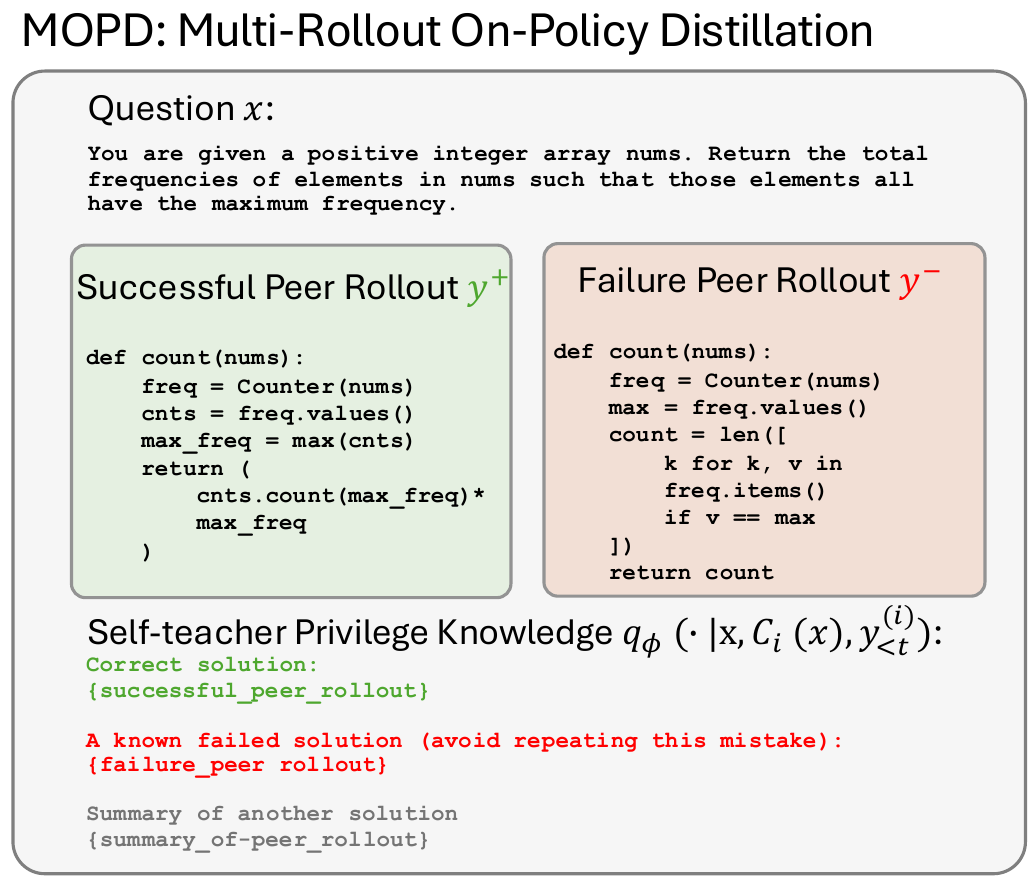}
    \vspace{-1.5em}
\caption{MOPD Illustration.} \label{fig:mopd}
\end{wrapfigure}

To directly examine whether peer conditioning improves the self-teacher signal itself, we introduce an analysis of self-teacher signal quality. For each prompt, we fix a set of student-generated rollouts containing both successful and failed attempts, vary only the context shown to the self-teacher, and compare the self-teacher’s normalized logits or scores with ground-truth verifier rewards. A better self-teacher signal should rank successful rollouts above failed ones, correlate more strongly with reward, and better separate correct from incorrect candidates. This analysis avoids the confound of answer-recovery tests and provides a direct measure of whether privileged contexts produce supervision that is actually aligned with correctness.

Empirically, MOPD improves over standard on-policy baselines across most scenarios on competitive programming, mathematical reasoning, scientific question answering, and tool-use benchmarks. The strongest gains arise from mixed peer contexts that combine successful and failed rollouts, supporting the hypothesis that contrastive evidence from peer rollout is more useful than positive demonstrations alone. Our self-teacher signal analysis further shows that the two-success-plus-one-failure context achieves the strongest alignment with verifier rewards across ranking and discrimination metrics, and our ablations show the same context yields the best compact downstream performance. These results indicate that effective distillation should exploit the student’s multi-rollout trial-and-error behavior, not only the strength of the teacher.

%% file: text/related_work_short.tex
\section{Related Work}
\label{sec:related-extended}

\paragraph{On-Policy Distillation}
Knowledge distillation transfers a teacher's predictive distribution to a
smaller student, but off-policy training creates a distribution mismatch:
the student is supervised on states it never visits at inference
\cite{hinton2015distilling, kim2016sequence, ross2011reduction}.
On-policy distillation addresses this by sampling trajectories from the
student's own policy and computing teacher supervision on those same
trajectories \cite{agarwal2024policy, 2306.08543, 2402.03898}.
Within this framework, work has focused on divergence objectives, token
reweighting, and interpolation between on- and off-policy regimes
\cite{2307.15190, 2510.24021, 2602.15260, 2603.07079}.
A parallel thread replaces an external teacher with self-distillation:
one copy of the model is conditioned on privileged context---solutions,
hints, or environment feedback---and supervises a second copy that lacks
it \cite{hubotter2024sdpo, ye2026policy, zhao2026self, 2602.04942, yang2026learning}.
Information can also be aggregated across multiple sampled solutions
through consistency, voting, or committee teachers
\cite{wangself, muennighoff2025s1, li2025mixtureteacher}.
However, the teacher signal for each rollout is
constructed independently, leaving cross-rollout structure within a
sampling group unexploited.

\paragraph{Multi-Rollout Reinforcement Learning and Verifier-Guided Improvement}
Post-training with verifiable rewards replaces learned reward models with
programmatic checkers and updates the policy from group-relative advantages
computed over multiple sampled rollouts per prompt
\cite{shao2024deepseekmath, wen2026RLVR, ouyang2022rlhf, rafailov2023dpo}.
Process reward models provide denser credit assignment by grading
intermediate reasoning steps \cite{cobbe2021training, setlur2025rewarding, zhang2025lessons},
and hybrid pipelines couple these signals with on-policy distillation
\cite{2505.16142, 2506.02208, 2602.22495}.
While these methods exploit multiple rollouts per prompt through advantage
normalization or rejection sampling, the supervisory signal for any single
trajectory is computed without conditioning on the contents of the others.
The work most adjacent to ours treats successes and failures as
complementary evidence at the level of advantages or filters, but stops
short of routing both into a single peer-conditioned distillation target. Extended related work for both topics is in \cref{appendix:related}.

%% file: text/prelim.tex
\section{Preliminaries}
\label{sec:preliminaries}

We consider the problem of distilling reasoning capabilities from a teacher model into a student model under an on-policy sampling regime. Let $x \in \mathcal{X}$ denote an input problem or prompt, and let
$y = (y_1,\ldots,y_T)$ denote an output trajectory generated autoregressively by a student policy
$\pi_\theta$. At decoding step $t$, we write the prefix as $y_{<t}$ and the token-level context as
$c_t=(x,y_{<t})$. The student distribution is $\pi_\theta(\cdot \mid c_t)$, while the teacher distribution is denoted by
$q_\phi(\cdot \mid c_t)$, possibly conditioned on additional privileged information.
In prior self-distillation methods, the privileged information available to the teacher typically includes ground-truth outcomes, verified successful answers \cite{ye2025black,ye2026policy,zhao2026self,yang2026learning}, or environment feedback \cite{hubotter2024sdpo}.

\paragraph{On-policy distillation.}
Offline distillation trains the student on trajectories sampled from a teacher or from a fixed dataset. This creates a distribution mismatch: during inference, the student may visit states that were not observed during training. On-policy distillation addresses this issue by first sampling trajectories from the student itself and then obtaining teacher supervision on those same trajectories. Formally, for each prompt $x$, the student generates $y \sim \pi_\theta(\cdot \mid x).$
The teacher then provides token-level targets along the student-induced trajectory. A standard on-policy distillation loss can be written as
\begin{equation}
    \mathcal{L}_{\mathrm{OPD}}
    =
    \mathbb{E}_{x \sim \mathcal{D},\, y \sim \pi_\theta(\cdot \mid x)}
    \left[
    \sum_{t=1}^{T}
    D\!\left(
        \pi_\theta(\cdot \mid x,y_{<t})
        \,\|\,
        q_\phi(\cdot \mid x,y_{<t})
    \right)
    \right],
\end{equation}
where $D(\cdot)$ denotes a divergence loss, either KL, reverse KL, or Jensen-Shannon (JS), as detailed in \cref{sec:divergence}. This formulation provides dense token-level supervision on the student's visited states. 

Nevertheless, standard OPD treats each sampled trajectory independently. When multiple rollouts are generated for the same prompt, the teacher distribution for one rollout is usually computed without access to the successes and failures observed in the other rollouts. This prevents the teacher from exploiting local, instance-specific evidence contained in the rollout group.


%% file: text/method.tex
\section{Multi-Rollout On-Policy Distillation}
\label{sec:method}

\begin{figure*}[t]
    \centering
    \includegraphics[width=0.9\textwidth]{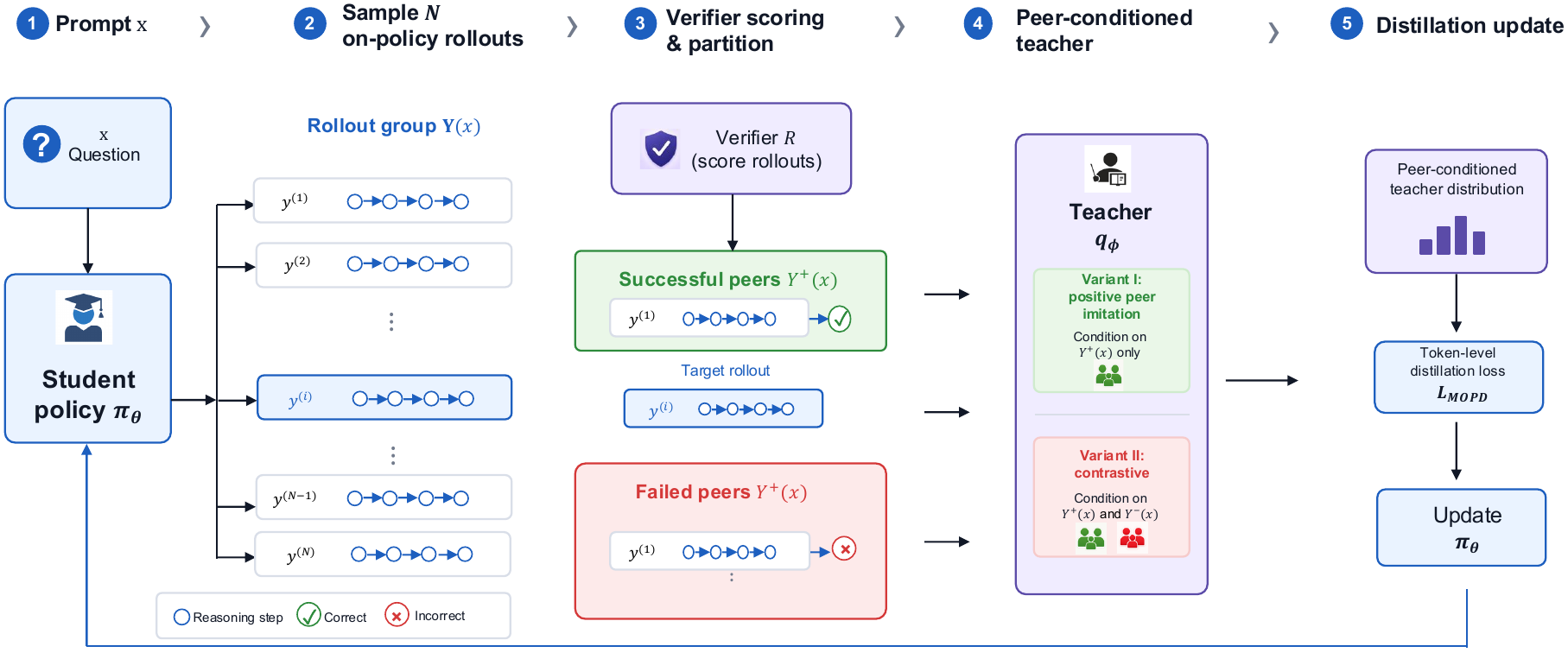} 
    \caption{MOPD Pipeline.}
    \label{fig:pipe}
\vspace{-1em}
\end{figure*}
We propose \textbf{Multi-Rollout On-Policy Distillation} (MOPD), a peer-conditioned distillation framework that exploits the local structure of multiple on-policy rollouts generated for the same question as in \cref{fig:pipe}. Standard on-policy distillation supervises each sampled trajectory independently: the teacher evaluates the student's current trajectory without direct access to the alternative attempts produced for the same instance. This ignores a useful source of instance-specific information. In reasoning tasks, successful rollouts reveal valid solution paths, while failed rollouts expose plausible but incorrect branches. MOPD uses this local contrast to construct a more informative teacher signal.

At a high level, MOPD first samples a group of student rollouts for each prompt, scores them with a verifier, partitions them into successful and failed trajectories, and then conditions the teacher on peer information from the same rollout group. The resulting teacher distribution is no longer only a global distribution over next tokens; it becomes a \emph{peer-conditioned} distribution that can compare the current trajectory against other local attempts. This allows the teacher to provide sharper token-level supervision, especially near branching points where failed trajectories diverge from successful ones.

\subsection{Multi-rollout peer conditioning}
\label{subsec:peer_conditioning}

For each prompt $x$, the student samples a group of $N$ on-policy trajectories,
\begin{equation}
\label{eq:rollout_group}
    Y(x)=\{y^{(i)}\}_{i=1}^{N},
    \qquad
    y^{(i)}\sim\pi_\theta(\cdot\mid x).
\end{equation}
Each trajectory is evaluated by a verifier or reward function $R$, and the rollout group is partitioned according to a success threshold $\tau$:
\begin{equation}
\label{eq:success_failure_partition}
    Y^+(x)=\{y^{(i)}:R(x,y^{(i)})\ge\tau\},
    \qquad
    Y^-(x)=\{y^{(i)}:R(x,y^{(i)})<\tau\}.
\end{equation}
Here $Y^+(x)$ contains peer successes and $Y^-(x)$ contains peer failures. For a target rollout $y^{(i)}$, we exclude the target itself and define
\begin{equation}
\label{eq:peer_sets}
    Y^+_{-i}(x)=Y^+(x)\setminus\{y^{(i)}\},
    \qquad
    Y^-_{-i}(x)=Y^-(x)\setminus\{y^{(i)}\}.
\end{equation}

MOPD constructs a peer context from these two sets:$
    C_i(x)
    =
    \mathcal{C}_v
    \left(
        x,
        y^{(i)},
        Y^+_{-i}(x),
        Y^-_{-i}(x)
    \right),$
where $\mathcal{C}_v$ denotes a context construction rule. The index $v$ specifies which peer-conditioning strategy is used. Given this context, the teacher distribution for the target trajectory becomes
$
    q_\phi^{\mathrm{peer}}(\cdot\mid x,y^{(i)}_{<t})
    =
    q_\phi(\cdot\mid x,C_i(x),y^{(i)}_{<t}).$
Compared with a standard teacher distribution $q_\phi(\cdot\mid x,y^{(i)}_{<t})$, the peer-conditioned teacher can use other rollouts from the same instance as local evidence. Successful peers provide positive support for correct reasoning patterns, while failed peers provide negative evidence about misleading or invalid trajectories.

The rollout-level MOPD loss is
\begin{equation}
\label{eq:mopd_rollout_loss}
    \mathcal{L}^{(i)}_{\mathrm{MOPD}}
    =
    \sum_{t=1}^{T_i}
    D\!\left(
        \pi_\theta(\cdot\mid x,y^{(i)}_{<t})
        \,\|\,
        q_\phi^{\mathrm{peer}}(\cdot\mid x,y^{(i)}_{<t})
    \right),
\end{equation}
where $D$ is a token-level divergence. In our implementation, $D$ is instantiated as reverse KL for math tasks, and Jensen--Shannon divergence for other tasks.


\subsection{Peer-context construction}
\label{subsec:peer_context_construction}

The main design choice in MOPD is the construction of $C_i(x)$. We compare two natural choices: a positive-only context that exposes the teacher to additional successful peers, and a contrastive context that adds failed peers as structured negative evidence. This comparison isolates whether failure-as-negative-evidence sharpens the teacher signal beyond success-only conditioning.

\input{text/alg}
\paragraph{Positive peer imitation.}
\label{subsubsec:positive_peer_imitation}

The first strategy conditions the teacher on successful peer trajectories only. Let $y^\star$ denote a primary successful rollout and let $y^+$ denote an additional successful peer, both selected from $Y^+_{-i}(x)$ when available. The context is
\begin{equation}
\label{eq:context_v1}
    C_i^{(1)}(x)
    =
    \mathrm{Template}(y^\star)
    \oplus
    \mathrm{SuccessTemplate}(y^+),
\end{equation}
where $\oplus$ denotes text concatenation. If only one successful peer is available, the context reduces to the primary successful trajectory:$C_i^{(1)}(x) =    \mathrm{Template}(y^\star).$

This variant treats successful rollouts as peer experts. It encourages the teacher to evaluate the target trajectory in light of alternative correct reasoning paths. Compared with using a single reference solution, positive peer imitation broadens the set of correct trajectories the teacher conditions on, so the distilled student is anchored to multiple valid derivation styles rather than one.

\paragraph{Contrastive success--failure conditioning.}
\label{subsubsec:contrastive_conditioning}

The second strategy conditions the teacher on both successful and failed peer trajectories. Given successful peers $y^\star,y^+\in Y^+_{-i}(x)$ and a failed peer $y^-\in Y^-_{-i}(x)$, we define
\vspace{-0.5em}
\begin{equation}
\label{eq:context_v2}
C_i^{(2)}(x) = \mathrm{Template}(y^\star) \oplus \mathrm{SuccessTemplate}(y^+) \oplus \mathrm{FailureTemplate}(y^-).
\end{equation}
\vspace{-1em}

This variant provides the teacher with contrastive local evidence. Successful rollouts indicate what the target trajectory should resemble, while failed rollouts identify plausible but incorrect reasoning patterns that should be avoided. The contrastive context can therefore sharpen token-level supervision by helping the teacher distinguish between superficially similar correct and incorrect trajectories. In tasks with sparse terminal rewards, this is especially useful because the failed peer exposes where an otherwise plausible solution path becomes invalid.

\paragraph{Unified view}

The two peer-context variants share a unified gated form: $C_i(x) = \mathrm{Template}(y^\star) \oplus \lambda_+ A_+(x) \oplus \lambda_- A_-(x)$, where $A_+(x)$ denotes additional successful demonstrations, $A_-(x)$ denotes failed demonstrations, and the gates $\lambda_+,\lambda_- \in \{0,1\}$ determine which type of peer evidence is included. Algorithm~\ref{alg:mopd} summarizes the MOPD training procedure.

%% file: text/alg.tex
\begin{wrapfigure}{r}{0.5\textwidth}
\vspace{-2em}
\begin{minipage}{0.5\textwidth}
\begin{algorithm}[H]
\small
\caption{MOPD Algorithm}
\label{alg:mopd}
\begin{algorithmic}[1]
\Require $\mathcal{D}$, $\pi_\theta$, $q_\phi$, $R$, $N$, $\tau$, $\mathcal{C}_v$, $\alpha$
\Ensure Updated student $\pi_\theta$
\For{each training iteration}
    \For{each $x \in \mathcal{B}$}
        \State Sample rollout group $Y(x)$ as in Eq.~\eqref{eq:rollout_group}.
        \State Score and partition rollouts as in Eq.~\eqref{eq:success_failure_partition}.
        \State Construct $C_i(x)$, query $q_\phi^{\mathrm{peer}}$, \\
           \Statex \qquad \qquad compute
               $\mathcal{L}^{(i)}_{\mathrm{MOPD}}$.
    \EndFor
    \State Update $\pi_\theta$ by minimizing \\
           \Statex \qquad $\mathcal{L}_{\mathrm{MOPD}} = \tfrac{1}{N}\sum_i \mathcal{L}^{(i)}_{\mathrm{MOPD}}$.
\EndFor
\State \Return $\pi_\theta$
\end{algorithmic}
\end{algorithm}
\end{minipage}
\vspace{-1em}
\end{wrapfigure}

%% file: text/exp_1.tex
\section{Experiments}
\label{sec:experiments}

We evaluate whether augmenting the teacher with peer rollouts improves post-training in reasoning-intensive domains. Our method is compared against the base model, GRPO, and SDPO. SDPO conditions its self-teacher on environment feedback when available; our method instead conditions the teacher on peer rollouts---both successful and failed---from the same prompt. This allows us to test whether peer-augmented teacher information provides additional benefits beyond scalar-reward RL and feedback-conditioned self-distillation.

\subsection{Experimental Setup}

\paragraph{Benchmarks.}
We consider four families of tasks that require multi-step reasoning or precise execution. \textbf{Science QA.}
We use the reasoning subsets (L3) from SciKnowEval~\citep{feng2024sciknoweval}, covering undergraduate-level questions in biology, chemistry, physics, and materials science.
\textbf{Tool Use.}
We evaluate tool-use reasoning on ToolAlpaca~\citep{tang2023toolalpaca}, where the model must map a user request and an API specification to the correct tool call.
\textbf{Code.}
We evaluate on LiveCodeBench v6 (LCBv6)~\citep{jain2025lcb}, which contains 131 contest-style programming problems released between February and May 2025. This split provides a recent and challenging evaluation of generalization to unseen coding problems.
\textbf{Math.}
For math training, we use a filtered subset of DeepMath~\citep{he2025deepmath}, selecting 57K examples with difficulty level at least 6 following~\citep{yang2026learning}. We evaluate on AIME2024, AIME2025, and HMMT25.
\paragraph{Metrics.}
For code and math, we report mean@8 and pass@8, following standard multi-sample evaluation. All results are reported as percentages.
\textbf{Training protocol.}
For each prompt, we sample a group of on-policy rollouts and evaluate them using a task-specific verifier or reward function. The resulting rollouts are partitioned into successful and failed sets. We then construct the privileged teacher context from peer rollouts, using two successful rollouts and one failed rollout unless otherwise specified. This design exposes the teacher to both correct solution patterns and informative failure modes. Across methods, we keep the rollout generation budget and optimization protocol fixed to ensure a fair comparison.

\textbf{Implementation details.}
Training is implemented in \texttt{verl} with asynchronous vLLM rollouts and FSDP-based optimization. We use a prompt batch size of 32 and sample 8 rollouts per prompt during training. The actor learning rate is set to $1\times 10^{-5}$ with 10 warmup steps. 
For generation, we use a maximum prompt length of 2048 tokens and a maximum response length of 4096 tokens (for math with 8192). Because SDPO conditions on additional solution context, we reserve an extra 2048-token headroom, yielding a total rollout context budget of 8192 tokens. Validation evaluates 8 samples per prompt. The configuration uses self-distillation with mixing coefficient $\alpha=0.5$ and \texttt{distillation\_topk}=100. We use FlashAttention-2 with bfloat16 model weights for both actor and critic. For training stability, we use reverse KL divergence for math tasks, and JS divergence otherwise.

\subsection{Main Results for Self-Distillation}

\begin{wrapfigure}{r}{0.4\textwidth}
    \centering
    \vspace{-1em}
    \includegraphics[width=0.4\textwidth]{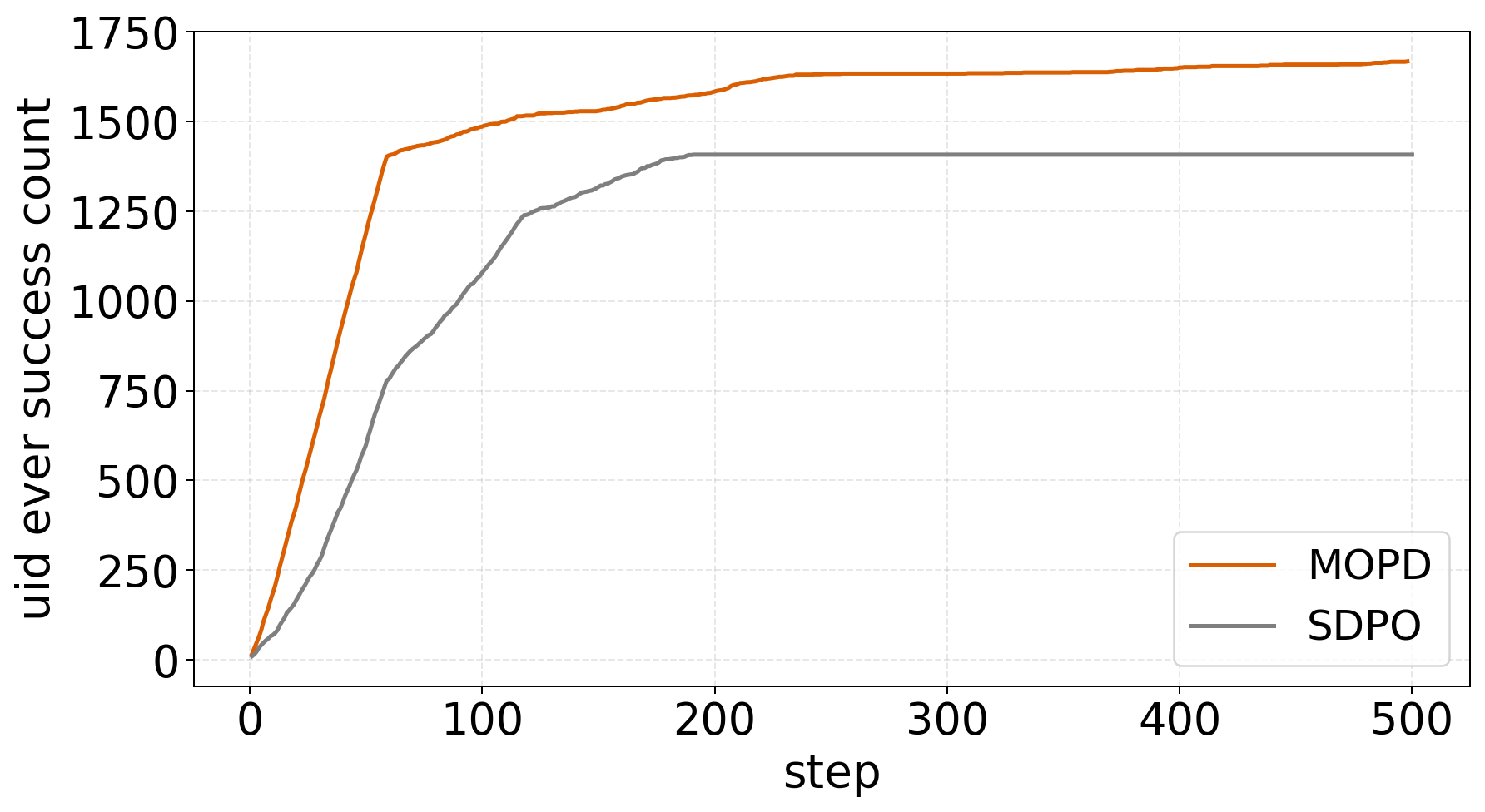} 
    \caption{Number of training data that have ever generated a correct answer in the $N$ rollout during training.}
    \label{fig:success_count}
    \vspace{-2em}
\end{wrapfigure}

\paragraph{LiveCodeBench.}
Table~\ref{tab:lcbv6} reports results on LCBv6 for Qwen3-4B and Qwen3-8B. LCBv6 provides structured environment feedback—such as compiler errors and test-case verdicts—at the end of each rollout. Our method consistently improves over the base model, GRPO, and SDPO at both model scales. On Qwen3-4B, our method improves over SDPO by large margin and on Qwen3-8B, our method reaches 61.82 mean@8 and 67.23 pass@8. 
\textbf{Math reasoning.}
Table~\ref{tab:math_main} shows results on AIME2024, AIME2025, and HMMT25 using Qwen3-4B as the base model. Our method substantially improves mean@8 compared to SDPO. {SDPO is designed for settings with rich environment feedback, but math tasks provide only a single binary reward signal from a verifier, which likely explains why it underperforms the baseline in this domain. }{Peer-augmented supervision is particularly effective for mathematical reasoning in self-distillation, where alternative successful derivations and failed attempts provide complementary evidence about solution structure that a single verifier reward cannot.}
\textbf{Science QA.}
Table~\ref{tab:scienceqa} shows MOPD achieves the best performance across all four domains, on science multi-choice questions benchmarks.
\textbf{Tool Use.}
Tool Use requires precise mapping from natural-language requests to structured API calls, our method outperforms. Peer conditioning therefore extends beyond reasoning-rich settings to tasks where the primary challenge is accurate structured-output prediction. \textbf{Time Cost.} MOPD introduces acceptable computational overhead over SDPO, with total per-step time increasing by only 9.1\% on LCB and 33.3\% on QA-Chemistry, detailed breakdown results in \cref{tab:compute}. \textbf{Number of ever-success questions during training.} As in \cref{fig:success_count}, MOPD reaches a higher count earlier and finally than SDPO, indicating faster exploration of the solution space across training prompts.

\paragraph{Case Study.} During training, we save the generated rollouts and compare them on the same question across training steps to provide a case study. Additionally, after training for the same number of steps, we save checkpoints from both SDPO and MOPD, then sample from these checkpoints to evaluate whether each model can successfully solve the target problem and avoid failure patterns. The details in \cref{sec:appendix-case-studies}, showing that \textbf{MOPD and SDPO start from the same failure pattern early in training, but by repeatedly exposing the teacher to failed peers of the same bug family and explicitly signaling 'avoid this mistake,' MOPD suppresses the error by checkpoint time} — while baseline method, lacking this channel, does not.

\begin{table}[t]
\centering
\caption{Math reasoning results on AIME2024/2025 and HMMT25. All methods start from the same Qwen3-4B model and use the same generation budget. Bold marks the best result in each column.}
\label{tab:math_main}
\setlength{\tabcolsep}{4pt}
{%
{\footnotesize
\begin{tabular}{l cc cc cc cc}
\toprule
& \multicolumn{2}{c}{\textbf{AIME2025}}
& \multicolumn{2}{c}{\textbf{AIME2024}}
& \multicolumn{2}{c}{\textbf{HMMT25 Feb.}}
& \multicolumn{2}{c}{\textbf{HMMT25 Nov.}} \\
\cmidrule(lr){2-3}
\cmidrule(lr){4-5}
\cmidrule(lr){6-7}
\cmidrule(lr){8-9}
\textbf{Method}
& mean@8 & pass@8
& mean@8 & pass@8
& mean@8 & pass@8
& mean@8 & pass@8 \\
\midrule
Qwen3-4B  
& 17.92 & 32.57 
& 16.89 & 31.60 
& 5.06 & 10.00 
& 9.58 & 13.91 \\
GRPO  
& \textbf{25.80} & \textbf{36.66} 
& 17.09 & 32.05
& 16.92 &  \textbf{20.94} 
& 13.16 & 19.53 \\
SDPO  
& 7.81 & 16.66 
& 7.29 & 16.29
& 8.59 &  13.61
& 12.01 & 18.49 \\
MOPD  
& {25.41} & {36.28} 
& \textbf{28.54} & \textbf{33.12} 
& \textbf{18.50} & {19.58} 
& \textbf{15.83} & \textbf{22.47} \\
\bottomrule
\end{tabular}}
}
\end{table}


%% file: text/exp.tex
\subsection{Analysis of Self-Teacher Signal Quality}

\begin{figure*}[t]
    \centering
    \includegraphics[width=\textwidth]{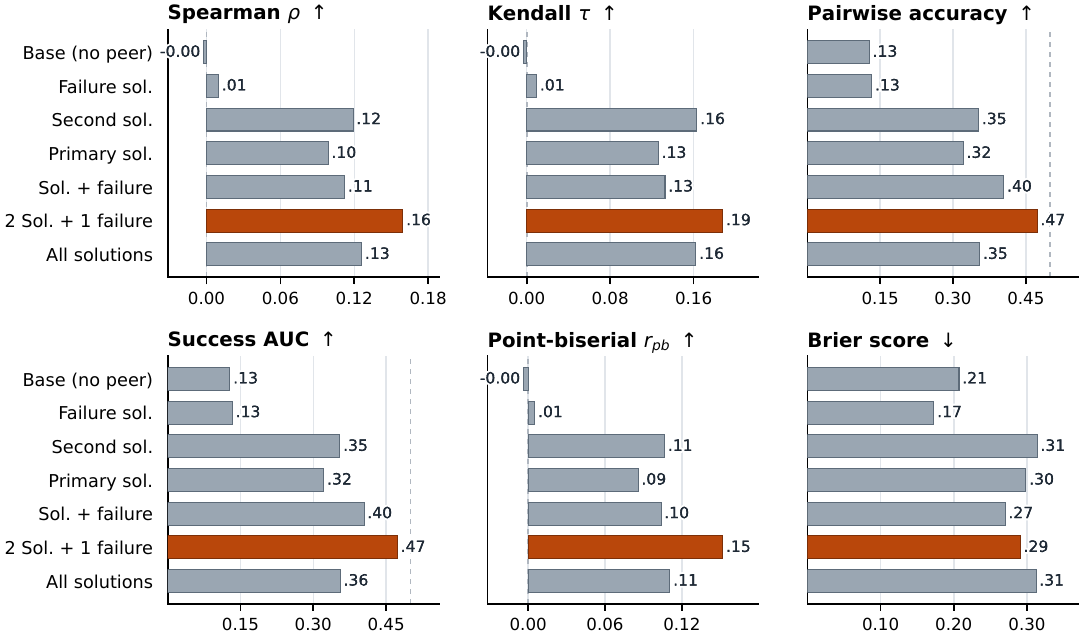} 
    \caption{Self-teacher-signal quality across seven context conditions. Each panel reports an averaged prompt-level metric. Higher is better for all metrics except the Brier score. 
}
    \label{fig:ana}
\vspace{-1em}
\end{figure*}

\begin{table}[t]
\centering
\begin{minipage}[t]{0.38\linewidth}
\centering
\caption{Science QA results.}
\label{tab:scienceqa}
{\footnotesize
\begin{tabular}{lcccc}
\toprule
\textbf{Method} 
& \textbf{Bio.} 
& \textbf{Chem.} 
& \textbf{Phys.} 
& \textbf{Mat.} \\
\midrule
Qwen3-8B & 30.89 & 42.98 & 58.44 & 65.59 \\
GRPO     & 47.32 & 64.24 & 64.12 & 72.09 \\
SDPO     & 50.60 & 62.91 & 67.36 & 72.34 \\
MOPD     & \textbf{55.69} & \textbf{74.29} & \textbf{76.25} & \textbf{78.59} \\
\bottomrule
\end{tabular}
}

\vspace{1em}

\caption{ToolUse results.}
\label{tab:tooluse}
{\footnotesize
\setlength{\tabcolsep}{3pt}
\begin{tabular}{lcccc}
\toprule
\textbf{Method} & Qwen3-8B & GRPO & SDPO & MOPD \\
\midrule
\textbf{Tool Use} & 59.11 & 60.32 & 63.45 & \textbf{66.73} \\
\bottomrule
\end{tabular}
}
\end{minipage}
\hfill
\begin{minipage}[t]{0.55\linewidth}
\centering
\caption{Ablation on peer-context construction for MOPD: which combination of successful and failed peer rollouts to inject into the self-teacher context.}
\label{tab:peer_context_ablation}
\setlength{\tabcolsep}{4pt}
\renewcommand{\arraystretch}{1.15}
{\footnotesize
\begin{tabular}{l cc cc}
\toprule
& \multicolumn{2}{c}{\textbf{QA-Chemistry}}
& \multicolumn{2}{c}{\textbf{LiveCodeBench}} \\
\cmidrule(lr){2-3}\cmidrule(lr){4-5}
\textbf{Peer context $\mathcal{C}(u)$}
& mean@8 & pass@8
& mean@8 & pass@8 \\
\midrule
1 successful solution         & 51.90 & 58.92 & 49.49 & 64.34 \\
2 successful solutions        & 52.62 & 59.02 & 55.62 & 64.02 \\
1 failure solution            & 35.67 & 37.64 & 46.72 & 54.86 \\
1 success + 1 failure         & 60.50 & 68.12 & 58.90 & 62.01 \\
2 success + 1 failure         & 74.29 & 83.53 & 61.82 & 67.23 \\
8 solutions (2 hours)         & 73.57 & 83.11 & 58.43 & 63.68 \\
8 solutions (4 hours)         & 81.25 & 84.49 & 58.95 & 64.71 \\
\bottomrule
\end{tabular}
}
\end{minipage}
\vspace{-1em}
\end{table}

\paragraph{Setup: self-teacher signal evaluation.}
To understand why peer information improves training, we directly analyze the quality of the self-teacher signal induced by different context constructions. For each question $x$, we sample a set of rollouts $\{y^{(i)}\}_{i=1}^{n}$ from the policy of the base model and evaluate each response using both the task verifier and the self-teacher model. Each response is associated with three quantities:
$(y^{(i)}, s_i, r_i),$
where $s_i \in \mathbb{R}$ is the self-teacher average token logprob $s_i = \frac{1}{T_i} \sum_{t=1}^{T_i} \log q_\phi\!\left(y_t^{(i)} \mid x,\, C_i(x),\, y_{<t}^{(i)}\right) $ assigned to response $(y^{(i)}$, $r_i \in [0,1]$ is the ground truth reward. The self-teacher score $s_i$ is computed under each specific teacher privilege knowledge, while $r_i$ is a fixed property of the candidate rollout. Thus, by holding the prompt and candidate responses fixed and varying only the teacher-visible context, we can isolate how different types of peer information affect the reliability of the self-teacher signal. We consider several context conditions, including base (no peer information), successful peer rollouts, failed peer rollouts, combinations of successful and failed rollouts, and all the rollouts. The analysis uses Qwen3-4B model generated on the LiveCodeBench-v6 benchmark.

We evaluate each self-teacher context condition with six complementary metrics covering ranking, discrimination, and calibration; full definitions are in \cref{appendix: metric}.
\textbf{Mean Spearman Correlation} and \textbf{Mean Kendall's $\tau$} both
measure ordinal agreement between teacher scores and ground-truth rewards
across rollouts.
\textbf{Pairwise Accuracy} directly reports the fraction of
reward-distinguishable rollout pairs that the teacher orders correctly. \textbf{Success AUC} measures the probability that a
successful rollout receives a higher teacher score than a failed one;
\textbf{Success Point-Biserial Correlation} quantifies the linear
association between teacher scores and binary success labels; and
\textbf{Success Brier Score (Sigmoid)} assesses probabilistic calibration
by mapping teacher scores through a sigmoid and computing MSE against rewards.

\begin{figure}[t]
\begin{minipage}[c]{0.45\linewidth}
\centering
\captionof{table}{LCBv6 results.}
\label{tab:lcbv6}
{\footnotesize
\begin{tabular}{lcccc}
\toprule
& \multicolumn{2}{c}{\textbf{Qwen3-4B}} 
& \multicolumn{2}{c}{\textbf{Qwen3-8B}} \\
\cmidrule(lr){2-3} \cmidrule(lr){4-5}
\textbf{Method} & \textbf{mean@8} & \textbf{pass@8} 
& \textbf{mean@8} & \textbf{pass@8} \\
\midrule
Base & 28.80 & 49.36 & 30.97 & 53.02 \\
GRPO & 40.75 & 55.43 & 43.65 & 58.72 \\
SDPO & 48.84 & 63.23 & 49.49 & 64.34 \\
Ours & \textbf{57.01} & \textbf{65.48} & \textbf{61.82} & \textbf{67.23} \\
\bottomrule
\end{tabular}
}
\end{minipage}
\hfill
\begin{minipage}[c]{0.5\linewidth}
\centering
\includegraphics[width=\linewidth]{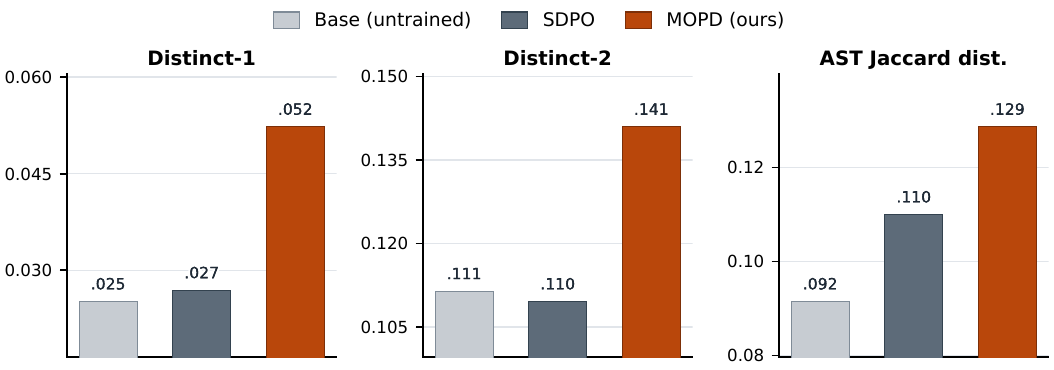}
\captionof{figure}{Diversity Analysis.}
\label{fig:diversity}
\end{minipage}
\end{figure}
\paragraph{Results and interpretation.}
Figure~\ref{fig:ana} and Table~\ref{tab:peer_context_ablation} together reveal how peer context composition affects teacher signal quality and downstream performance. 1) Successful peer rollouts are the primary driver of teacher-signal quality: any condition containing at least one successful peer lifts pairwise accuracy from $0.13$ at the no-peer baseline to above $0.32$. A failed peer alone barely moves the discrimination metrics. 2) Successful peers have a stronger effect on correctness-related metrics---Spearman $\rho$ rises from 0.00 to above 0.10 and Kendall $\tau$ from 0.00 to above 0.13 when at least one successful solution is included---suggesting that seeing a correct peer helps the teacher better assess the target rollout's quality. 3) Failed peers complement successful ones: adding a failure to a success-only context lifts pairwise accuracy from $0.32$ to $0.40$, showing that negative evidence sharpens decision boundaries that positive evidence alone leaves blurred. 4) Combining both types yields the best results: the ``2 success + 1 failure'' context achieves the highest score on $5$ of the $6$ ranking and discrimination metrics in the signal-quality analysis, with a competitive Brier score, and the highest LCB downstream mean@$8$ among the compact peer-context settings. 5) We notice that the results for base are very low, for example, the pairwise accuracy and success AUC is only 0.13. We look into the generations of Qwen3-4B and find that \textbf{57\%} of prompts have no successful rollout 
($Y^+(x) = \emptyset$), explaining the near-zero metrics in the hard bucket. 
This confirms that correct solutions occupy extremely low-probability regions 
of the model's output distribution---the self-teacher finds them genuinely 
surprising, which directly explains why model accuracy remains below 30\% on 
lcb-v6. 6) On LiveCodeBench, providing all available rollouts yields no additional gains beyond the compact two-success-plus-one-failure context; more peer rollouts do not strictly dominate a balanced contrastive pair on every benchmark.

%% file: text/exp_3.tex
\subsection{Solution Diversity Analysis}
\label{sec:diversity}

We investigate whether MOPD's exposure to multiple peer rollouts during training leads to greater diversity in generated solutions. We measure two axes of diversity that are well defined for code: Distinct-1 and Distinct-2 for within-response lexical richness (the vocabulary and bigram coverage inside a single program), and AST-node Jaccard distance for between-response structural variation (the difference between programs in their parsed syntax-tree composition). Both are computed on LiveCodeBench-v6 checkpoints at step 50, with each model generating 8 responses per problem. Detailed implementation and metric definitions are provided in \cref{appendix:diversity}.

Figure~\ref{fig:diversity} shows that MOPD produces substantially more diverse outputs than both the base model and SDPO across all three metrics. On Distinct-1 and Distinct-2, SDPO remains close to the untrained base model (0.027 vs.\ 0.025; 0.110 vs.\ 0.111), suggesting that its training signal does not encourage lexical variation. MOPD, by contrast, achieves nearly double the Distinct-1 score of SDPO (0.052 vs.\ 0.027) and a notable gain in Distinct-2 (0.141 vs.\ 0.110). A similar pattern holds at the structural level: AST Jaccard distance increases from 0.110 (SDPO) to 0.129 (MOPD), indicating that MOPD generates solutions with more structurally distinct implementations. Conditioning the teacher on failed peer rollouts therefore improves both correctness and the lexical and structural diversity of the student's outputs.

\subsection{Does Peer Information Improve Teacher--Student Distillation?}
\label{sec:rq3_teacher_student}

We next study whether peer information remains useful in a teacher--student setting, where the supervision signal is provided by a larger teacher model and used to train a smaller student. Our general knowledge distillation (GKD) pipeline follows an offline teacher-forcing paradigm: for each input prompt, a larger teacher model generates 8 completions, which are then converted into supervised fine-tuning examples. The student is trained by maximum-likelihood learning on these teacher responses.
In on-policy distillation (OPD), the teacher provides supervision conditioned on on-policy student responses. Ours TS method further augments this teacher context with peer information: successful and failed peer rollouts. 

\begin{table}[th]
\centering
\vspace{-0.5em}
\caption{Teacher--student on-policy distillation results.
All methods share the same rollout budget and hyperparameter.}
\label{tab:ts_distillation}
\setlength{\tabcolsep}{8pt}
\renewcommand{\arraystretch}{1.15}
\begin{tabular}{l cc cc}
\toprule
& \multicolumn{2}{c}{\textbf{ToolUse}}
& \multicolumn{2}{c}{\textbf{LiveCodeBench}} \\
\cmidrule(lr){2-3}\cmidrule(lr){4-5}
\textbf{Method}
& mean@8 & pass@8
& mean@8 & pass@8 \\
\midrule
Qwen3-4B (S)       & 53.27 & 58.58 & 28.80 & 49.36 \\
Qwen3-14B (T)      & 56.02 & 60.89 & 46.75 & 65.16 \\
GKD (TS) & 55.91 & 60.05 & 46.05 & 56.67 \\
OPD (TS) & 59.63 & 61.58 & 60.92 & 67.30 \\
SDPO (Self)     & 62.04 & 65.97 & 48.84 & 63.23 \\
\textbf{Ours (TS)}   & \textbf{66.44} & \textbf{68.61} & \textbf{61.92} & \textbf{67.35} \\
\textbf{Ours (Self)} & {64.61} & {66.81} & 57.01 & 65.48 \\
\bottomrule
\end{tabular}
\end{table}

Table~\ref{tab:ts_distillation} reveals several trends in peer-conditioned distillation. 1) On ToolUse and LiveCodeBench, traditional teacher-student setups with on-policy strategies still outperform self-distillation.  2) When the teacher is given richer context through peer information, distillation can surpass the teacher model's own performance---a result that underscores the value of privileged observations at inference time. 3) Adding peer information consistently improves both the self-distillation and teacher--student settings, confirming that the benefit is not specific to a particular pairing of model sizes. 4) On-policy methods outperform their off-policy counterparts across both domains, consistent with the distribution-shift argument motivating on-policy training.

%% file: text/conclusion.tex
\section{Conclusion}

We introduced MOPD, a peer-conditioned on-policy distillation framework that uses multiple student rollouts from the same prompt to construct more informative teacher signals. By conditioning the teacher on successful and failed peer rollouts, MOPD turns independent trajectory supervision into a comparative learning process: successes provide positive evidence for valid reasoning, while failures expose error patterns to avoid. 
Experiments across code, math, science QA, and tool use show that MOPD consistently improves over standard on-policy baselines. Both ablations and teacher-signal analysis indicate that mixed success–failure contexts produce the most faithful supervision, better aligning teacher scores with verifier rewards and downstream performance. Effective distillation should exploit the student’s rollout group as a structured collection of successes and failures, rather than treating each trajectory in isolation.

%% file: text/appendix.tex
\section{Teacher Signal Quality Metric}
\label{appendix: metric}
\paragraph{Mean Spearman Correlation.}
Spearman correlation measures the agreement between the rank ordering induced by teacher scores and the rank ordering induced by rewards. For a prompt, let $\mathrm{rank}(s_i)$ and $\mathrm{rank}(r_i)$ denote the average ranks under ties. The prompt-level Spearman correlation is
\[
\rho_{\mathrm{S}} \;=\; \mathrm{corr}\!\bigl(\mathrm{rank}(s_1),\dots,\mathrm{rank}(s_n),\;
\mathrm{rank}(r_1),\dots,\mathrm{rank}(r_n)\bigr),
\]
where $\mathrm{corr}(\cdot,\cdot)$ is the Pearson correlation coefficient. We report the mean of this quantity across prompts. Higher values indicate that the teacher better preserves the reward-induced ordering.

\paragraph{Mean Kendall's $\tau$.}
Kendall's $\tau$ evaluates pairwise ranking consistency between teacher scores and rewards. For each unordered pair $(i,j)$ with $i<j$, let the pair be \emph{concordant} if $(s_i-s_j)(r_i-r_j)>0$ and \emph{discordant} if $(s_i-s_j)(r_i-r_j)<0$. Pairs tied in either score or reward are ignored. The prompt-level metric is
\[
\tau \;=\; \frac{C-D}{C+D},
\]
where $C$ and $D$ are the numbers of concordant and discordant pairs, respectively. We then average $\tau$ across prompts. Higher values indicate stronger ordinal agreement.

\paragraph{Pairwise Accuracy.}
Pairwise accuracy measures the fraction of unequal-reward pairs that are ordered correctly by the teacher. Formally,
\[
\mathrm{PairAcc}
\;=\;
\frac{
\sum_{i<j} \mathbf{1}[r_i \neq r_j]\mathbf{1}\!\left[(s_i-s_j)(r_i-r_j)>0\right]
}{
\sum_{i<j} \mathbf{1}[r_i \neq r_j]
}.
\]
Unlike Kendall's $\tau$, this metric does not penalize incorrect pairs symmetrically around zero; instead, it directly reports the proportion of reward-distinguishable pairs ranked correctly. Higher is better.

\paragraph{Success AUC.}
Success AUC treats the teacher score as a binary classifier score for success. Let $\mathcal{P}=\{i:y_i=1\}$ and $\mathcal{N}=\{j:y_j=0\}$. The prompt-level AUC is computed by comparing all positive--negative pairs:
\[
\mathrm{AUC}
\;=\;
\frac{1}{|\mathcal{P}||\mathcal{N}|}
\sum_{i \in \mathcal{P}} \sum_{j \in \mathcal{N}}
\left(
\mathbf{1}[s_i > s_j] + \frac{1}{2}\mathbf{1}[s_i=s_j]
\right).
\]
This is the probability that a randomly chosen successful response receives a higher teacher score than a randomly chosen unsuccessful one, with ties receiving half credit. Higher is better.

\paragraph{Success Point-Biserial Correlation.}
To quantify the linear association between teacher scores and binary success labels, we compute the point-biserial correlation, which is simply the Pearson correlation between the continuous teacher scores and the binary labels:
\[
\rho_{\mathrm{PB}} \;=\; \mathrm{corr}(s_1,\dots,s_n,\; r_1,\dots,r_n).
\]
A larger positive value indicates that successful responses tend to receive higher teacher scores.

\paragraph{Success Brier Score (Sigmoid).}
To assess probabilistic calibration of teacher scores with respect to success, we first map each score to a pseudo-probability using the sigmoid transform
\[
p_i \;=\; \sigma(s_i) \;=\; \frac{1}{1+e^{-s_i}},
\]
and then compute the Brier score
\[
\mathrm{Brier}
\;=\;
\frac{1}{n}\sum_{i=1}^n (p_i - r_i)^2.
\]
This metric measures the mean squared error between predicted success probabilities and observed binary outcomes. Lower values indicate better calibration.

\begin{table*}[t]
\centering
\small
\setlength{\tabcolsep}{4pt}
\begin{tabular}{lcccccc}
\toprule
\textbf{Condition} 
& \textbf{Mean} & \textbf{Mean} & \textbf{Pairwise} & \textbf{Success} & \textbf{Success} & \textbf{Success} \\
& \textbf{Spearman} & \textbf{Kendall's $\tau$} & \textbf{Acc.} & \textbf{AUC} & \textbf{Point-Biserial} & \textbf{Brier (Sig.)} \\
\midrule
Base &-0.0024   & -0.0031  &0.1275   &0.1283  &-0.0037  &0.2071 \\
Primary Sol.       & 0.0988 & 0.1262 & 0.3211 & 0.3217 & 0.0859 & 0.2986 \\
Second Sol.        & 0.1190 & 0.1632 & 0.3528 & 0.3547 & 0.1065 & 0.3139 \\
Failure Sol.       & 0.0097 & 0.0095 & 0.1326 & 0.1332 & 0.0052 & 0.1728 \\
Sol.+Failure       & 0.1122 & 0.1326 & 0.4045 & 0.4045 & 0.1038 & 0.2701 \\
2 Suc.+1 Failure   & 0.1595 & 0.1879 & 0.4733 & 0.4733 & 0.1520 & 0.2908 \\
All Solutions      & 0.1260 & 0.1623 & 0.3550 & 0.3556 & 0.1107 & 0.3124 \\
\bottomrule
\end{tabular}
\caption{Prompt-level offline teacher-signal metrics averaged over prompts for the six
context conditions.
Higher is better for all metrics except Success Brier (Sigmoid), where lower is better.}
\label{tab:teacher-signal-metrics-appendix}
\end{table*}

Table~\ref{tab:teacher-signal-metrics-appendix} reports prompt-level teacher-signal 
metrics across the six context conditions. The \textit{2 Suc.+1 Failure} condition 
consistently achieves the strongest performance, attaining the highest Mean Spearman 
($0.1595$), Mean Kendall's $\tau$ ($0.1879$), Pairwise Accuracy ($0.4733$), Success AUC 
($0.4733$), and Success Point-Biserial ($0.1520$), indicating that a mixed context of two 
successful and one failed solution provides the richest teacher signal. In contrast, the 
\textit{Failure Sol.} condition performs the worst across nearly all ranking and 
discrimination metrics, with a Mean Spearman of only $0.0097$ and a Pairwise Accuracy of 
$0.1326$, suggesting that failure demonstrations alone offer little useful signal. 
\textit{Failure Sol.} attains the lowest Brier score ($0.1728$), but its discrimination metrics also collapse to near-zero, so this calibration value reflects a degenerate near-uniform output rather than a useful signal. Among the success-only conditions, \textit{Second Sol.} 
slightly outperforms \textit{Primary Sol.} and \textit{All Solutions} across most metrics, 
while \textit{Sol.+Failure} improves pairwise accuracy over either success-only condition alone.

\section{Preliminary on Divergence-based distillation}
\label{sec:divergence}

\paragraph{Divergence-based distillation.}
A common approach to distillation is to minimize a divergence between the teacher distribution
$q_\phi$ and the student distribution $\pi_\theta$ at each token position. In language model distillation, the direction of the KL divergence induces qualitatively different behavior.

The \emph{forward KL} objective,
\begin{equation}
    D_{\mathrm{KL}}\!\left(q_\phi(\cdot \mid c_t)
    \,\|\, \pi_\theta(\cdot \mid c_t)\right),
\end{equation}
penalizes the student for assigning low probability to tokens favored by the teacher. It is therefore mode-covering: when the teacher assigns mass to multiple plausible continuations, forward KL encourages the student to preserve this diversity.

The \emph{reverse KL} objective,
\begin{equation}
    D_{\mathrm{KL}}\!\left(\pi_\theta(\cdot \mid c_t)
    \,\|\, q_\phi(\cdot \mid c_t)\right),
\end{equation}
penalizes the student for placing probability mass on tokens that the teacher considers unlikely. This objective is mode-seeking and is often attractive in on-policy settings because the expectation is taken over the student distribution. However, reverse KL may over-concentrate probability mass on a subset of teacher-preferred modes, especially when the teacher distribution is uncertain or when the supervision signal is incomplete.

A symmetric alternative is the Jensen--Shannon divergence,
\begin{equation}
    D_{\mathrm{JS}}(p \,\|\, q)
    =
    \frac{1}{2}D_{\mathrm{KL}}(p \,\|\, m)
    +
    \frac{1}{2}D_{\mathrm{KL}}(q \,\|\, m),
    \qquad
    m = \frac{1}{2}(p+q).
\end{equation}
$D_{\mathrm{JS}}$ is symmetric and bounded. In principle, it can provide a more stable compromise between mode-covering and mode-seeking behavior. In practice, however, computing such distribution-level objectives over the full vocabulary may be expensive for large language models, motivating approximations based on sampled tokens, top-$k$ support, or teacher-provided log-probabilities.

\input{text/case_study}

\section{Diversity}
\label{appendix:diversity}




\paragraph{Distinct-$n$.}
Distinct-$n$~\citep{li2016diversity} is the ratio of unique $n$-grams to total
$n$-grams across the concatenated token stream of all responses:
\begin{equation}
  \mathrm{Distinct\text{-}}n = \frac{|\{\text{unique }n\text{-grams}\}|}
                                     {\text{total }n\text{-grams}}.
\end{equation}
We report Distinct-1 and Distinct-2.


\paragraph{AST node Jaccard distance (code only).}
Each response is parsed into a Python AST and its node-type multiset
$\mathbf{a}_i$ is extracted.
Pairwise Jaccard distance is
\begin{equation}
  d^\mathrm{AST}_{ij} =
    1 - \frac{\sum_t \min(a_{it},\, a_{jt})}
             {\sum_t \max(a_{it},\, a_{jt})},
\end{equation}
where $t$ ranges over all AST node types present in either response.
Responses that fail to parse are excluded; the metric is omitted for a prompt
if fewer than two responses parse successfully.
The per-prompt score is the mean over all valid pairs.

\paragraph{Code Extraction for Diversity Measurement.}
For code-focused diversity analysis, we do not compute metrics on the raw response text directly. Instead, each response is first converted into a code view using a simple fenced-code-block extraction rule. Specifically, we match Markdown code blocks of the form \texttt{```python ... ```} or \texttt{``` ... ```} with a regular expression, and if multiple code blocks appear in one response, we keep only the last matched block. The extracted block is then stripped of leading and trailing whitespace and used as the canonical code content for subsequent diversity computation. If no fenced code block is found, the entire response text is used as a fallback. This procedure is intentionally lightweight: it does not perform full Markdown parsing, and only the \texttt{python} language tag is explicitly removed. As a result, responses with other language tags may retain the tag text inside the extracted content.

\paragraph{Results.}
Table~\ref{tab:diversity} summarises the diversity metrics
across the three candidate sets evaluated on {131} shared prompts with
up to eight rollouts each.

\begin{table}[h]
\centering
\caption{Diversity metrics (mean over prompts, $n{=}131$).
         Higher is better for all columns.
         \textbf{Bold} marks the best value per column.}
\label{tab:diversity}
\begin{tabular}{lccc}
\toprule
Method & Distinct-1 & Distinct-2 & \makecell{AST Jaccard\\dist.} \\
\midrule
MOPD     & \textbf{0.052} & \textbf{0.141} & \textbf{0.129} \\
SDPO     & 0.027 & 0.110 & 0.110 \\
Baseline & 0.025 & 0.111 & 0.092 \\
\bottomrule
\end{tabular}
\end{table}

MOPD produces rollouts with substantially higher lexical and structural
diversity: Distinct-1 is roughly $2{\times}$ that of SDPO and the baseline,
and the mean pairwise AST node-type Jaccard distance is {17}\% higher
than SDPO.

\section{More Training Details and Training Statistics}
\paragraph{Compute Resources.}
All self-distillation experiments are conducted on a single node of 8$\times$H100 GPUs and a single node of 8$\times$A100 GPUs. Teacher--student distillation experiments are conducted on 2 nodes each equipped with 8$\times$H100 GPUs.

\begin{table}[h]
\centering
\caption{Per-step teacher-query cost comparison. Teacher context length is measured in tokens
(mean over training steps). Wall-clock time is measured per gradient update step on 8$\times$H100 GPUs
with batch size 32 and 8 rollouts per prompt.}
\label{tab:compute}
\begin{tabular}{llccccc}
\toprule
Dataset & Method & Time Gen & Time Reward & Time Adv & Time Update & Total \\
\midrule
\multirow{2}{*}{LCB}          & SDPO & 37 & 400 & 1 & 78  & 550 \\
                               & MOPD & 37 & 400 & 2 & 90  & 600 \\
\midrule
\multirow{2}{*}{QA-Chemistry}  & SDPO & 5  & 0.1 & 1 & 75  & 90  \\
                               & MOPD & 5  & 0.1 & 2 & 110 & 120 \\
\bottomrule
\end{tabular}
\end{table}

\paragraph{Time Cost.}
We report several wall-clock timing metrics in seconds to break down the cost of each training iteration. To avoid startup overhead, we report the average timing measured after the first five training steps, since earlier steps may include additional one-time loading and initialization costs. \texttt{timing\_s/gen} measures the time spent generating sampled responses from the current policy. \texttt{timing\_s/reward} measures the time required to obtain training rewards, including reward-model scoring or task-specific reward computation. \texttt{timing\_s/update\_actor} measures the wall-clock time of the actor optimization stage. In the SDPO setting, this stage includes the student forward pass, the teacher forward pass used to produce teacher logits, the self-distillation loss computation, and the subsequent backward pass and optimizer update.
Finally, \texttt{timing\_s/adv} measures the time spent in the advantage-processing stage, which includes reward post-processing and the computation of training advantages used for policy optimization. \texttt{timing\_s/step} measures the total duration of one training step, covering the full pipeline from rollout generation to optimization. The reported stage times do not sum exactly to the full step time because they only cover the main timed components of the training loop. The total step time also includes additional overhead such as data movement, batch reorganization, mask construction, metric logging, coordination between workers, and other small bookkeeping operations that are not exposed as separate timing entries. As a result, the difference between the sum of the listed stage times and the total step time can be understood as uncategorized system and orchestration overhead.

Table~\ref{tab:compute} reports the per-step wall-clock time breakdown for SDPO and MOPD across two datasets. The dominant cost in both methods is reward computation and parameter update; rollout generation and advantage estimation are comparatively negligible. MOPD introduces additional overhead primarily in the update phase, as the teacher must process longer peer-conditioned contexts. Despite this, the total per-step time increases modestly: from 550s to 600s on LCB \textbf{(a 9.1\% increase)} and from 90s to 120s on QA-Chemistry \textbf{(a 33.3\% increase)}. These results indicate that the computational overhead of peer conditioning is small relative to the overall training cost, and that MOPD's consistent performance gains are not simply an artifact of increased compute budget.

\begin{table}[h]
\centering
\caption{Science QA results with standard deviation over 3 runs. Bold marks the best result in each column.}
\label{tab:sciqa_errbar}
\begin{tabular}{lcccc}
\toprule
Method & Biology & Chemistry & Physics & Materials  \\
\midrule
SDPO     & $50.60_{\pm 2.1}$ & $73.22_{\pm 2.5}$ & $67.36_{\pm 1.9}$ & $72.34_{\pm 2.6}$  \\
MOPD       
    & ${55.69}_{\pm 1.6}$ & ${74.29}_{\pm 2.0}$ & ${76.25}_{\pm 2.8}$ & ${78.59}_{\pm 1.9}$  \\
\bottomrule
\end{tabular}
\end{table}

\paragraph{Training stability.} Table~\ref{tab:sciqa_errbar} reports the mean and standard deviation of MOPD across 3 independent 
runs on the Science QA benchmarks. While MOPD consistently achieves the best performance across 
all four domains, the standard deviations reveal moderate variance, particularly in Physics 
($\pm 2.8$) and Chemistry ($\pm 2.0$), suggesting that training is not fully stabilized across 
runs. We attribute this to the stochastic nature of on-policy rollout sampling and the sensitivity 
of peer-context construction to the specific success/failure partition observed in each run. 
Improving training stability through more robust peer selection or variance reduction techniques 
remains an avenue for future work.

\input{text/related_new}

\input{text/appendix_more}

%% file: text/case_study.tex
\section{Case Studies}
\label{sec:appendix-case-studies}

We present three complementary forms of evidence for MOPD's qualitative
advantage.
Case Studies 1 and 2 isolate checkpoint-level semantic failures: both models
see the same task, but MOPD preserves the crucial constraints that SDPO drops.
Case Study 1 additionally provides direct training-time evidence linking the
checkpoint gap to MOPD's peer-conditioning mechanism.
Case Study 3 examines training dynamics, showing that MOPD internalizes correct
behavior earlier while SDPO remains unstable late in training.

\subsection{Case Study 1: Identifier-space composition
            (checkpoint gap + training-time mechanism)}

\paragraph{Candidate pair.}
The bib-mapping task provides a direct SDPO-versus-MOPD comparison.
On this prompt, the SDPO checkpoint at training step~50 reaches accuracy $0.175$, while the MOPD checkpoint at the same step solves it perfectly.

\paragraph{SDPO failure.}
The SDPO trajectory mixes up the role of the permutation arrays:
\begin{verbatim}
bib_to_person = [0] * (N + 1)
for i in range(N):
    bib_to_person[Q[i]] = P[i]
result = [str(bib_to_person[Q[i]]) for i in range(N)]
\end{verbatim}
This is incorrect in two ways: \texttt{P[i]} is the stared-at person, not the
wearer of bib \texttt{Q[i]}, and the final answer stops in person-ID space
instead of projecting back to the target person's bib number.
The program preserves the rough shape of the mapping task but composes the
wrong relations.

\paragraph{MOPD success.}
The corresponding MOPD sample keeps the two identifier spaces separate
throughout:
\begin{verbatim}
bib_to_person = [0] * (N + 1)
for i in range(N):
    bib_to_person[Q[i]] = i + 1
for i in range(1, N + 1):
    person = bib_to_person[i]
    target_person = P[person - 1]
    result.append(str(Q[target_person - 1]))
\end{verbatim}
This trajectory performs the full chain correctly:
\texttt{bib} $\rightarrow$ \texttt{wearing person}
$\rightarrow$ \texttt{stared-at person}
$\rightarrow$ \texttt{target bib}.

\paragraph{Training-time context: MOPD encounters the same failure pattern
during rollout.}
The same bib-mapping prompt appears in the MOPD rollout stream at training step~1 with seven failed peers and one successful peer; the failed peers exhibit the same identifier-space ambiguity that SDPO retains at checkpoint time. Because MOPD injects failed peers directly into the teacher context, the teacher supervising the successful target on this prompt is exposed to a concrete instance of this bug paired with an explicit ``avoid this mistake'' instruction. SDPO has no equivalent channel.

\subsection{Case Study 2: Dropped middle-character constraint}

\paragraph{Candidate pair.}
The evenly spaced \texttt{A}-\texttt{B}-\texttt{C} counting task gives a
second clean contrast.
On this prompt, the SDPO checkpoint at training step~50 reaches accuracy $0.825$, while the MOPD checkpoint at the same step is fully correct.

\paragraph{SDPO failure.}
The SDPO code enforces equal spacing and checks the two endpoints, but drops
the requirement that the middle character must be \texttt{B}:
\begin{verbatim}
for j in range(1, n-1):
    for i in range(j):
        if S[i] == 'A':
            k = 2 * j - i
            if k < n and S[k] == 'C':
                count += 1
\end{verbatim}
This overcounts invalid triples such as \texttt{A-R-C} in the test case
\texttt{ARC}, where the correct answer is $0$.
The failure is a local semantic omission: the positional structure is
preserved, but one label constraint from the specification is forgotten.

\paragraph{MOPD success.}
The MOPD sample reinstates the missing check explicitly:
\begin{verbatim}
for j in range(1, n - 1):
    if s[j] == 'B':
        for i in range(j):
            if s[i] == 'A':
                k = 2 * j - i
                if k < n and s[k] == 'C':
                    count += 1
\end{verbatim}
The program is structurally close to the SDPO attempt but preserves the full
symbolic constraint set.

\paragraph{Takeaway.}
Case Studies 1 and 2 together show that MOPD's advantage in avoiding the failure pattern is not tied to a single error type.
In Case Study 1, SDPO fails by composing the wrong identifier relations; here
it fails by dropping one clause from the specification.
MOPD succeeds in both settings; the contrast holds across two distinct error families rather than depending on a single failure mode.

\subsection{Case Study 3: Training dynamics ---
            MOPD stabilizes early, SDPO fails late}

\paragraph{Candidate pair.}
The direction-opposite task illustrates the training-time gap.
The SDPO checkpoint fails on this prompt at both step~50 and step~100 with accuracy $0.0$: the step-50 sample is truncated, and the step-100 sample is flagged for a missing \texttt{```python} block. The MOPD checkpoint at step~50 already solves the same prompt with accuracy $1.0$.

\paragraph{Early MOPD signal.}
In the MOPD rollout stream, a correct solution first appears at rollout step~10 with score $1.0$---the
earliest observed point at which the repaired behavior is present in
training-time rollouts for this prompt.

\paragraph{Late SDPO failure.}
At rollout step~80,
SDPO still fails with accuracy $0.0$, receiving feedback:
\texttt{Incorrect Format: Put your code inside a ```python ... ``` block.}

\paragraph{Takeaway.}
The gap here is not algorithmic: MOPD internalizes basic output-format
discipline by approximately step~10, while SDPO has not stabilized that
behavior even late in training.

\subsection{Overall Pattern}

Taken together, the three case studies in lcv benchmark reveal a consistent picture.
The baseline's failures are narrow and consequential---a dropped constraint, a confused
identifier space, an unresolved format requirement---rather than wholesale
algorithmic breakdowns.
These are precisely the errors that peer conditioning is designed to surface:
by exposing the teacher to failed peers that exhibit the same bug, MOPD
produces a sharper supervision signal at the critical decision points.
Case Study~1 compared through training logs;
Case Studies 2 and 3 provide corroborating evidence across a different failure
type and a training-dynamics lens.
The combined evidence supports the conclusion that MOPD's gains arise from a
more targeted teacher signal, not from discovering fundamentally different
solution strategies.

%% file: text/related_new.tex
\section{More Related Work}
\label{appendix:related}

\paragraph{On-Policy Distillation of Large Language Models}
Knowledge distillation classically transfers a teacher's predictive distribution to a smaller student through supervision on teacher- or human-curated sequences \cite{hinton2015distilling, kim2016sequence, 2402.13116}. For autoregressive language models, this off-policy regime exhibits a well-known limitation: the student is trained on states it never visits at inference, and prediction errors compound over long generations \cite{2305.15717, ross2011reduction, song2026surveyonpolicydistillationlarge}. The central response is to move the supervision signal onto trajectories drawn from the student's own evolving policy, anchoring training on the distribution that will be operated over at deployment \cite{agarwal2024policy, 2306.08543, 2402.03898, 2503.07067}.
Within this framework, a substantial body of work focuses on how to construct and weight the per-token learning signal. Forward-, reverse-, and skewed-divergence objectives offer different trade-offs between mode coverage and mode seeking \cite{2306.08543, 2402.03898, 2307.15190, 2404.02657, 2505.16297}. Other works adapt the supervision to vocabulary mismatches between teacher and student \cite{2402.12030}, reweight tokens by uncertainty or salience \cite{2510.24021, 2503.02832}, or interpolate between fully on- and off-policy regimes via teacher-prefixed rollouts \cite{2410.11325, 2602.15260}. Black-box variants drop the teacher-logit requirement and supervise the student through sampled outputs, sequence-level rewards, or adversarial discrimination \cite{ye2025black, 2305.12870}. The paradigm now underlies widely deployed post-training pipelines for instruction-following, code generation, and mathematical reasoning \cite{2505.09388}.
A complementary thread documents stability and failure modes of these objectives, including capacity-induced learnability gaps, divergence pathologies, and entropy collapse, and proposes adaptive losses or curricula to mitigate them \cite{2502.08606, li2026rethinkingonpolicydistillationlarge, 2603.07079}. Despite this breadth, the dominant assumption is that the teacher signal for each rollout is constructed independently, leaving cross-rollout structure within a sampling group unexploited.

\paragraph{Self-Distillation and Self-Improvement for Reasoning}
A second line of work treats the same model as both teacher and student, exploiting either privileged conditioning or repeated sampling to manufacture supervision without an external teacher. Iterative self-training procedures generate candidate solutions, filter by verification or reward, and fine-tune on the surviving trajectories, yielding a coarse form of self-distillation for reasoning and alignment \cite{zelikman2022star, gulcehre2023rest, singh2023restem, chen2024self, yang2024self}. Subsequent work develops soft self-distillation: one copy of the model is conditioned on additional context---ground-truth solutions, hints, demonstrations, or environment feedback---and produces per-token targets for a second copy that lacks that conditioning \cite{ye2026policy, zhao2026self, hubotter2024sdpo, 2601.19897, 2602.04942, 2603.23871, yang2026learning}.
Self-distillation has been particularly active in long chain-of-thought reasoning, where supervision is sparse and individual traces are expensive. One direction compresses or reshapes long reasoning traces during distillation \cite{2603.05433, 2603.11137}. Another aggregates information across multiple sampled solutions through consistency, voting, or committee teachers, transforming an ensemble of attempts into a richer learning signal \cite{wangself, muennighoff2025s1, li2025mixtureteacher, snell2024scaling}. A third direction interleaves rollouts with structured self-critique or search before constructing supervision, expanding the teacher's effective view of the problem \cite{zhang2024rest-mcts, bao2025fixing, zhang2025critiquegrpo}. Recent analyses also document the limitations of these procedures, noting that self-distillation can degrade reasoning when the student is asked to imitate signals beyond its competence or when the privileged context misaligns with what the student can reproduce \cite{2603.24472}. Across these directions, the common thread is that a single model can sharpen its own training signal when given structured access to additional information or to alternative attempts; whether and how those alternative attempts should be jointly conditioned on within a single distillation step remains underexplored.

\paragraph{Multi-Rollout Reinforcement Learning and Verifier-Guided Improvement}
A third line of related work concerns post-training paradigms that exploit groups of rollouts and external verification rather than per-token teacher matching. Reinforcement learning from human or AI feedback aligns model behavior with preference labels collected over pairs or groups of completions, with subsequent advances exploring offline, reference-free, and online variants \cite{ouyang2022rlhf, rafailov2023dpo, hong2024orpo, meng2024simpo, casper2023open}. Reinforcement learning with verifiable rewards extends this idea by replacing learned reward models with programmatic checkers---answer matching, unit tests, type-checkers---and updating the policy from group-relative advantages computed over multiple sampled rollouts per prompt \cite{shao2024deepseekmath, wen2026RLVR, wu2025support, yue2025doesrl, fu2025srft, hipo2026}. A unified perspective on off- and on-policy post-training places these methods alongside on-policy distillation and clarifies when each provides the most useful supervision signal \cite{zhao2026largelanguagemodelposttraining, 2504.14945, 2512.23097}.
Process- and outcome-level verifiers refine the supervision beyond a single sparse trajectory reward. Step-level verifiers and process reward models grade intermediate states, providing denser credit assignment for reasoning \cite{2305.20050, cobbe2021training, setlur2025rewarding, zhang2025lessons, guan2024verifier, dixit2026aletheia}. Hybrid post-training pipelines couple on-policy distillation with reinforcement learning from verifiers, using teacher-derived signals as an auxiliary objective alongside reward optimization \cite{2505.16142, 2506.02208, 2509.14257, 2602.22495}. While these methods do leverage multiple rollouts per prompt---typically through advantage normalization, rejection sampling, or success-conditioned filtering---the supervisory signal for any single trajectory is computed without conditioning on the contents of the others. The work most adjacent to ours therefore treats successes and failures from the same group as complementary streams of evidence at the level of advantages or filters, but stops short of routing both into a single, peer-conditioned distillation target.

%% file: text/appendix_more.tex
\section{Limitation}
MOPD relies on the availability of multiple rollouts and a verifier or reward function that can partition them into successful and failed trajectories. This makes the method most directly applicable to verifiable domains such as code, math, tool use, and structured QA, while extending it to open-ended generation tasks may require reliable preference models or process-level evaluators. Peer conditioning increases context length and teacher-query cost when raw rollouts are concatenated, especially for long reasoning traces; selecting the most informative peers under context-length constraints is an open question. Finally, the quality of the teacher signal depends on the correctness of the success/failure labels: noisy verifiers or mislabeled rollouts may introduce misleading positive or negative evidence. Future work should study more robust peer selection, better handling of noisy feedback, and extensions to less easily verifiable tasks.

\section{Societal Impacts}

MOPD may have positive societal impacts by improving the efficiency of post-training and making stronger reasoning capabilities more accessible in smaller language models. Better distillation can reduce inference cost and energy use, and may help deploy capable models in resource-constrained settings. In domains such as code generation, tool use, scientific question answering, and mathematical reasoning, more reliable small models could support education, research assistance, and productivity tools.

At the same time, improving the reasoning and tool-use capabilities of smaller models may also increase risks. More capable and cheaper models could be misused for generating harmful code, automating deceptive content, or executing tool-use workflows in unintended ways. Because MOPD learns from both successful and failed trajectories, poor verifier design or biased feedback could also reinforce undesirable behavior if incorrect success/failure labels are treated as reliable supervision. These risks suggest that peer-conditioned distillation should be paired with careful verifier design, safety evaluations, misuse monitoring, and appropriate deployment safeguards, especially when applied to open-ended, high-impact, or tool-using systems.

\section{Declaration of LLM Usage}

We used large language models as assistance tools during the preparation of this paper. Specifically, LLMs were used to help draft, revise, and polish parts of the manuscript, including improving clarity, grammar, and organization. We also used LLM assistance for code writing and debugging, such as drafting utility scripts, checking implementation details, and improving code readability.